\documentclass[letterpaper, 10 pt, conference]{ieeeconf}  
\IEEEoverridecommandlockouts
\title{
    \LARGE \bf UCON: Uncertainty-aware Navigation with \\ Historical Re-association in Dynamic Environments
}

\author{
    Bing Sun, Yue Lin, Yongsheng Yuan, Yang Liu, Dong Wang*, Huchuan Lu
    \thanks{
        All authors are with Dalian University of Technology, Dalian 116024.China. *Corresponding author: Dong Wang, wdice@dlut.edu.cn.
    }
    \thanks{
        This work is supported by the National Natural Science Foundation of China (Nos. U23A20384, 62293542, and 62476044), the Dalian Outstanding Science and Technology Talent Project (No. 2025RJ01), the Liaoning Science and Technology Joint Program Project (No. 2024011188-JH2/1026), the Open Research Fund from Guangdong Laboratory of Artificial Intelligence and Digital Economy (SZ) (No. GML-KF-24-19), as well as the Open Research Fund from National Key Laboratory of China Space Intelligent Control Technology (No. HTKJ2024KL502016).
    }
}

\usepackage{graphicx}
\usepackage{amsmath}
\usepackage{amssymb}
\usepackage{booktabs}
\usepackage{multirow}
\usepackage{tabularx}
\usepackage{makecell}
\usepackage{array}
\usepackage{cite}

\usepackage[linesnumbered, ruled]{algorithm2e}
\makeatletter
\renewcommand{\@algocf@pre@ruled}
\makeatother

\begin{document}

\thispagestyle{empty}
\pagestyle{empty}

\maketitle

\begin{abstract}
    Autonomous navigation in dynamic environments is hindered by two fundamental challenges: perception instability and uncertainty–optimization mismatch. The former leads to identity switches and unreliable motion estimation, while the latter prevents principled incorporation of motion uncertainty into trajectory optimization. To address these challenges, we propose UCON, an uncertainty-aware navigation algorithm in dynamic environments. For perception instability, we present a point-level historical re-association mechanism that leverages historical point cloud fragments to recover lost targets while maintaining identity continuity. Subsequently, a Kalman filter is employed to provide anisotropic motion state estimation and covariance propagation. To resolve the uncertainty–optimization mismatch, we transform predicted states and their covariances into uncertainty sectors, which are embedded as differentiable cost terms within a trajectory optimization framework. This achieves consistent uncertainty-aware dynamic obstacle avoidance while maintaining smoothness and feasibility. Extensive simulations and real-world experiments demonstrate that, while maintaining high computational efficiency, UCON achieves superior perception stability and robust navigation performance in dynamic environments compared to state-of-the-art methods. The code will be open-sourced to facilitate further research.
\end{abstract}

\section{Introduction}

Autonomous navigation is a fundamental robotic task that enables safe and efficient operation in complex environments. Recent research has achieved significant progress in geometrically complex scenarios~\cite{zhou2019robust, zhou2020ego, 10924228, wang2025fast}. However, dynamic settings introduce additional challenges that extend beyond geometric complexity. In these scenarios, autonomous navigation is hindered by two critical issues:
\begin{itemize}
    \item \textit{Perception instability}. Dynamic objects undergoing occlusion, clustering, or partial observation can lead to identity switches and inconsistent motion estimation.
    \item \textit{Uncertainty–optimization mismatch}. Motion uncertainty cannot be consistently incorporated into trajectory optimization, resulting in overly conservative or insufficiently safe behaviors.
\end{itemize}

Existing studies have attempted to mitigate perception instability. Event camera-based approaches~\cite{falanga2020dynamic, Guan_2025_ICCV, zhu2023event} provide promising alternatives but are often cost-prohibitive for practical robotics. Traditional vision-based methods~\cite{xu2023real, 11246819, xu2023onboard, 11180066, zhang2026observability} are susceptible to motion blur and illumination variations. In contrast, LiDARs are widely adopted due to their robust and high-precision geometric sensing capabilities across diverse lighting conditions and ranges. However, existing LiDAR-based methods~\cite{wu2024moving, xu2025flow, lu2024fapp, zhu2024swarm} predominantly rely on object-level association, which can lead to perception instability in cases of partial occlusion or sparse point clouds. Furthermore, from a planning perspective, these methods optimize trajectories based solely on average predicted states and fixed safety margins, neglecting uncertainty-optimization mismatch. Therefore, existing systems have not yet established a principled and tightly coupled mechanism that bridges perception uncertainty and trajectory optimization. This structural decoupling between perception and planning fundamentally limits navigation performance in dense dynamic environments.

\begin{figure}[t]
    \vspace{2 mm}
    \centering
    \includegraphics[width=\linewidth]{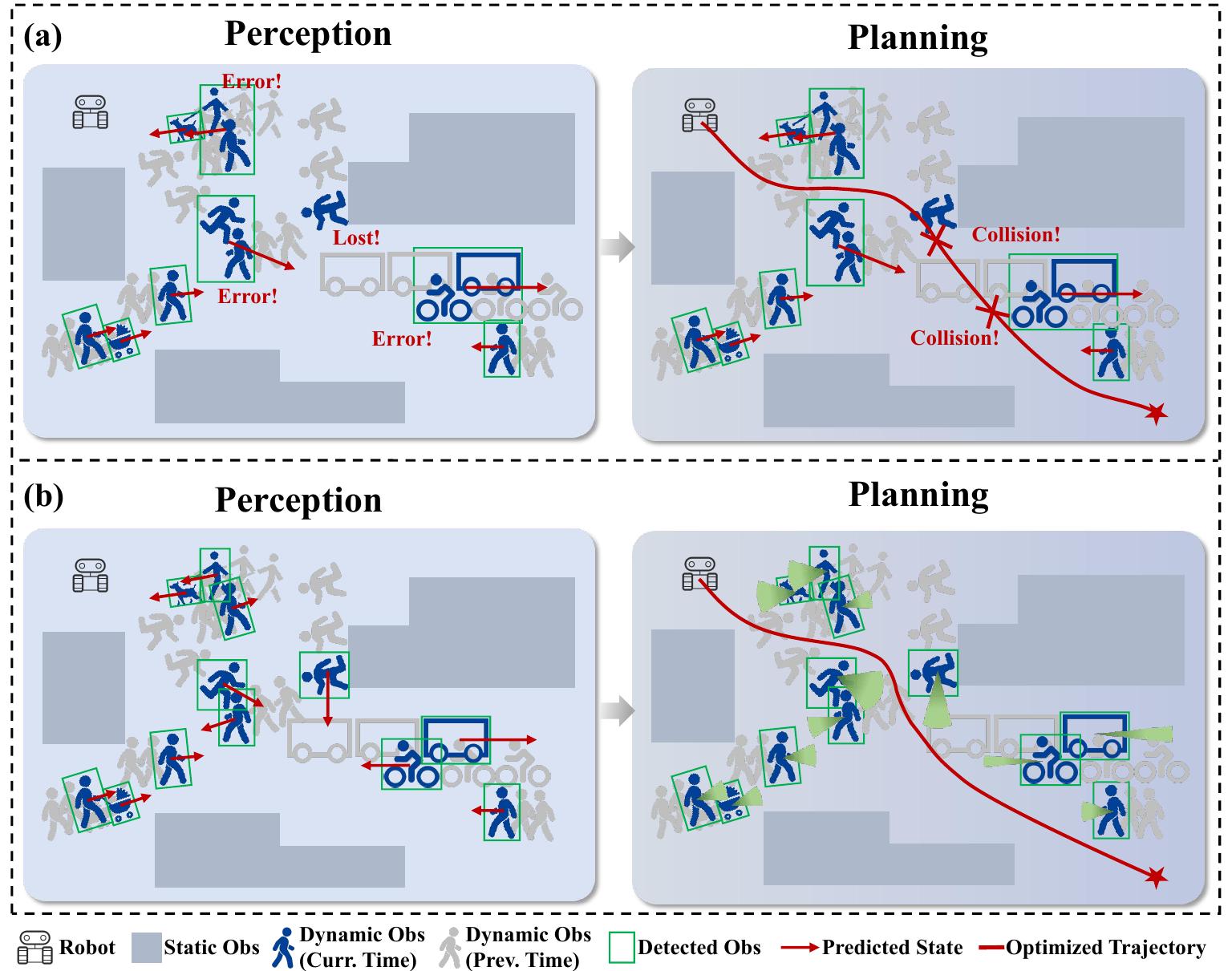}
    \caption{Comparison between the existing methods and the proposed UCON. (a) Limitations of existing dynamic navigation approaches. (b) The proposed UCON maintains identity continuity via historical re-association and embeds anisotropic motion uncertainty in trajectory optimization.}
    \vspace{-4 mm}
    \label{fig:head}
\end{figure}

\begin{figure*}[t]
    \vspace{2 mm}
    \centering
    \includegraphics[width=\linewidth]{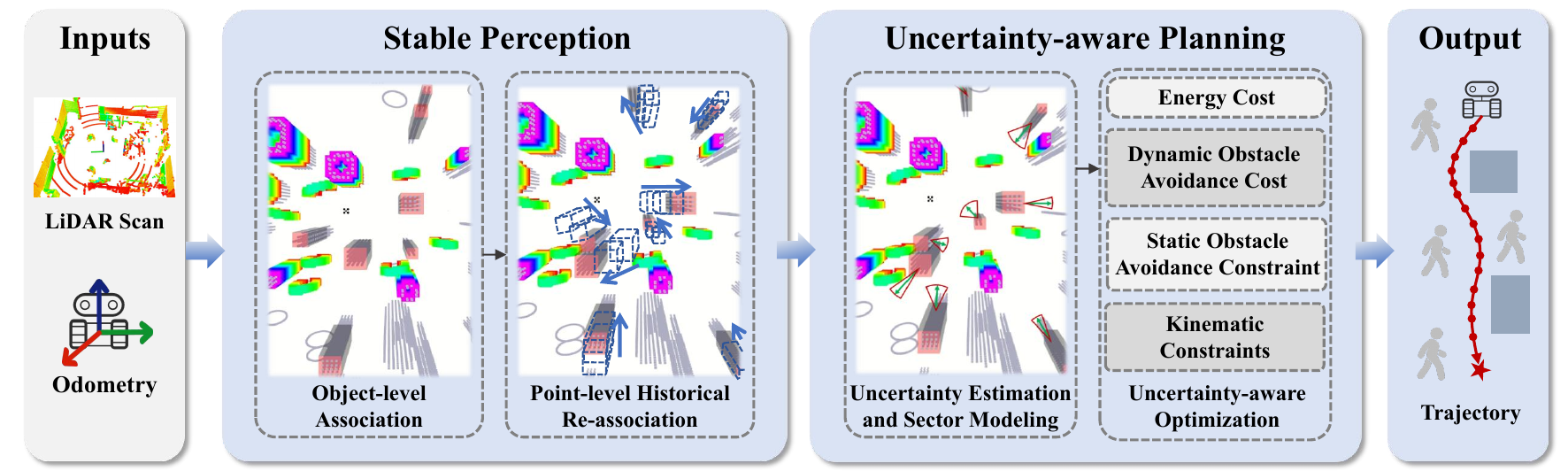}
    \caption{Overview of the proposed UCON. UCON integrates stable perception with uncertainty-aware trajectory optimization. Taking LiDAR point clouds and robot pose as inputs, the system first distinguishes between dynamic and static obstacles. The stable perception module initially employs an object-level association mechanism to detect dynamic objects, followed by a point-level historical re-association mechanism to recover lost objects, thereby ensuring perception stability. The uncertainty-aware planning module estimates the motion prediction covariance of dynamic objects and generates uncertainty sectors. During the trajectory optimization stage, these sectors are explicitly converted into differentiable cost functions for dynamic obstacle avoidance, ultimately producing a smooth and safe trajectory.}
    \vspace{-3 mm}
    \label{fig:framework}
\end{figure*}

To address the aforementioned challenges, we propose UCON, an \textbf{U}n\textbf{C}ertainty-aware navigati\textbf{ON} algorithm. As illustrated in Fig.~\ref{fig:head}, unlike previous methods, UCON simultaneously mitigates perception instability and resolves the uncertainty-optimization mismatch. To alleviate perception instability, UCON introduces a point-level historical re-association mechanism that leverages historical point cloud fragments to recover lost objects and maintain identity continuity under occlusion and clustering. By reconstructing motion states from temporally consistent observations, it improves perception stability. Subsequently, a Kalman filtering framework provides motion state estimation along with anisotropic covariance propagation. To address the uncertainty–optimization mismatch, UCON transforms predicted obstacle states and their covariances into uncertainty sectors. These sectors explicitly encode anisotropic motion uncertainty and are embedded as differentiable cost terms within a trajectory optimization framework. This modeling strategy enables principled and uncertainty-aware obstacle avoidance. By tightly coupling stable perception with differentiable and uncertainty-aware optimization, UCON establishes a consistent pipeline from dynamic state estimation to trajectory generation, as shown in Fig.~\ref{fig:framework}.

To validate the proposed UCON, we conduct extensive simulations and real-world experiments in dynamic cluttered environments. Experimental results demonstrate that UCON significantly improves perception stability, achieving an accuracy of 85.2\% with an average processing latency of 17.78 ms, while markedly reducing identity switches compared to state-of-the-art methods~\cite{lu2024fapp,xu2025intent}. Furthermore, in dense dynamic scenarios with up to 110 moving obstacles, UCON consistently achieves higher navigation success rates without introducing noticeable computational or energy overhead. Finally, experiments conducted in real-world dynamic environments further demonstrated the practicality of UCON.

In summary, the major contributions of this paper are:
\begin{itemize}
    \item A historical re-association mechanism to mitigate the perception instability.
    \item An uncertainty-aware sector modeling strategy to resolve the uncertainty–optimization mismatch.
    \item A tightly coupled perception–planning pipeline for navigation in dynamic environments, supported by extensive experimental validation.
\end{itemize}

\section{Related Work}

\subsection{Attempts to
Mitigate Perception Instability}

Stable perception is a prerequisite for reliable navigation in dynamic environments. With the advancement of neural networks, learning-based methods~\cite{ravipati2024object, tabernik2024center, ghanta2025space} have demonstrated satisfactory perception accuracy. However, their substantial computational overhead prevents these methods from achieving real-time inference on low-power embedded devices of robotic systems. Mainstream LiDAR-based perception methods~\cite{wu2024moving, xu2025flow, lu2024fapp, zhu2024swarm} adopt object-level association strategies that depend on bounding box overlap or centroid proximity. However, these approaches are highly vulnerable to identity switches or target loss when encountering sparse point clouds or overlapping object clusters, which can subsequently lead to perception failures. Therefore, maintaining identity consistency to alleviate perception instability under deteriorating observation conditions remains an unsolved problem.

\subsection{Uncertainty–optimization Mismatch in Planning}

Trajectory planning in dynamic environments has been extensively studied. Classical velocity-obstacle methods~\cite{fiorini1998motion, chen2022real, han2022reinforcement} provide collision constraints but typically assume deterministic obstacle motion, which is often unrealistic. Model predictive control~\cite{xu2025intent, xu2022dpmpc} is widely utilized for obstacle avoidance in dynamic environments. However, this method only optimizes the control quantity in the local time domain, making it difficult to achieve global trajectory optimality and easily getting trapped in local optima. Learning-based approaches~\cite{loquercio2021learning, xie2023drl, tordesillas2023deep} attempt to model interactive behaviors and multi-modal predictions. Although these methods improve prediction realism, uncertainty is frequently treated implicitly or approximated through fixed safety margins. Optimization-based planners~\cite{wang2021autonomous, lu2022perception, zhang2025threat} incorporate predicted trajectories but often rely on isotropic safety inflation to compensate for uncertainty. Even when probabilistic motion estimates are available, the associated covariance information is rarely consistently integrated into trajectory optimization. Therefore, trajectory planning is conducted under simplified or mismatched representations of uncertainty. Consequently, a structural gap persists between probabilistic perception outputs and deterministic trajectory optimization frameworks. 

\section{Mitigating Perception Instability}

To mitigate dynamic perception instability, we design a two-stage data association framework. It integrates object-level association and point-level historical re-association.

\subsection{Object-level Association}

Following the previous work FAPP~\cite{lu2024fapp}, dynamic objects are first extracted from raw LiDAR point clouds through geometric clustering and motion-based filtering. In the data association phase, unlike the FAPP, which is based on a single metric, we employ a two-stage association strategy. First, we establish high-confidence state continuity through object-level association, and then recover targets lost through a point-level historical re-association mechanism.

Let $C$ denote the set of objects detected at the current time, and $H$ denote the set of objects tracked in the previous frame. We model object-level association as a bipartite graph matching problem, with $C$ and $H$ as the vertex sets. The cost function between $c \in C$ and $h \in H$ is defined as
\begin{equation}
    f(c, h) = \mu_1(1 - u(c, h)) + \mu_2(1 - \hat{\mathbf{v}}_c \hat{\mathbf{v}}_h) + \mu_3\dfrac{\Vert \boldsymbol{\delta}(c,h)\Vert}{\Vert \mathbf{s}_h \Vert},
\end{equation}
where $u(c, h)$ represents the intersection-over-union ratio of the projections of the bounding boxes of $c$ and $h$ onto the $x$-$y$ plane, $\hat{\mathbf{v}}_i$ denotes the unit motion direction vector of object $i$, $\boldsymbol{\delta}(c, h) = \mathbf{s}_c - \mathbf{s}_h$, $\mathbf{s}_i$ denotes the size vector of the projection of the bounding box of object $i$ onto the $x$-$y$ plane, and $\mu_1$, $\mu_2$, and $\mu_3$ are hyperparameters used to balance the various terms. Besides, we project the bounding boxes onto the $x$-$y$ plane during the data association stage. This dimensionality reduction effectively filters out noise from sensor observations along the $z$-axis.

\subsection{Point-level Historical Re-association}

The object-level association becomes fragile in cluttered environments. When objects partially occlude one another or experience temporary detection loss, the matching process may incorrectly assign identities or prematurely terminate trajectories. These errors accumulate over time, leading to identity switches and fragmented motion histories. This instability arises because object-level matching treats each frame independently and relies solely on aggregated geometric representations. The absence of fine-grained temporal memory hinders recovery after association failures, thereby directly contributing to perception instability.

To mitigate perception instability, we explicitly leverage historical point cloud fragments to restore identity continuity, as illustrated in Fig.~\ref{fig:reassociation}. The detailed process of point-level historical re-association algorithm is outlined in Appendix~\ref{app:algorithm}. Specifically, let $\hat C$ denote the set of objects in $C$ that were not successfully matched by the object-level association. For each object $c \in \hat C$, we extract its point cloud set $T_c$, and let $N_c$ represent the number of points in $T_c$. Then, we construct a K-D tree for the current environment point cloud set $P$ combined with the point clouds of all unmatched objects in $\hat C$. For each object $c \in \hat C$, we traverse the points $\mathbf{p} \in T_c$ and use the K-D tree to find the nearest point $\mathbf{q}$ to $\mathbf{p}$ in $P$, calculating the distance
\begin{equation}
    d = \Vert \mathbf{p} + \mathbf{v}_c \tau - \mathbf{q} \Vert,
\end{equation}
where $\tau$ is the time interval between two observations, and $\mathbf{v}_c$ is the predicted velocity of $c$. If $d$ is less than a preset threshold $d_0$, we determine that point $\mathbf{p}$ is successfully matched. After traversing all points, if the number of successfully matched points exceeds $\gamma N_c$, we update the point cloud of object $c$ with the successfully matched points in $P$, where $\gamma \in (0, 1)$ is a hyperparameter.

\begin{figure}[t]
    \centering
    \vspace{2 mm}
    \includegraphics[width=\linewidth]{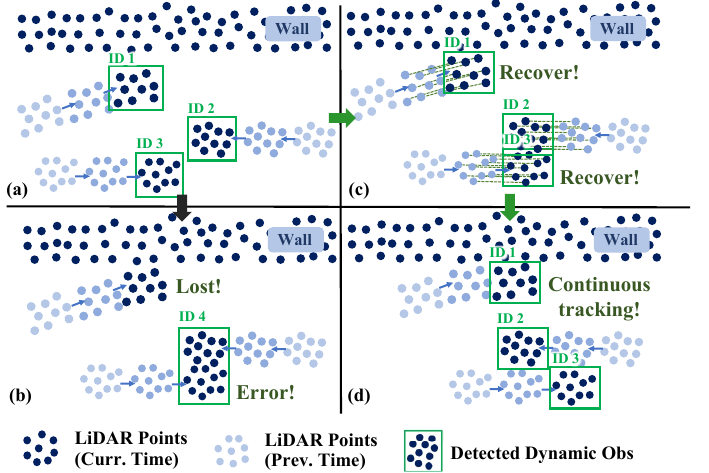}
    \caption{Illustration of point-level historical re-association. (a) The perception module needs to identify three dynamic objects and a static wall. (b) Traditional object-level association methods suffer from perception instability when dynamic objects are close to static objects or when two dynamic objects are close to each other. (c) The proposed point-level historical re-association mechanism re-associates points based on historical information. (d) The three dynamic objects are successfully detected, and their identities remain consistent.}
    \label{fig:reassociation}
    \vspace{-3 mm}
\end{figure}

\section{Uncertainty–aware Trajectory Optimization}

\subsection{Uncertainty Estimation}

To provide uncertainty-aware inputs to the trajectory optimization module, we adopt a Kalman filter for recursive motion state estimation. The state of an object is defined as $\mathbf{x}_t = [\mathbf{p}_t^\top, \mathbf{v}_t^\top]^\top \in \mathbb{R}^6$, where $\mathbf{p}_t, \mathbf{v}_t \in \mathbb{R}^3$ denote the position and velocity at time step $t$, respectively. The system model of each object can be formulated as
\begin{equation}
    \mathbf{x}_{t+1} = \mathbf{F} \mathbf{x}_t + \mathbf{w}_t, \quad
    \mathbf{F} = \begin{bmatrix}
        \mathbf{I} & \tau \mathbf{I} \\
        \mathbf{O} & \mathbf{I}
    \end{bmatrix},
\end{equation}
where $\mathbf{I}$ and $\mathbf{O}$ represent the identity matrix and the zero matrix, respectively, and $\mathbf{w}_t \sim \mathcal{N}(\mathbf{0}, \mathbf{Q})$ is the normally distributed process noise with covariance matrix $\mathbf{Q}$, which characterizes the incremental prediction risk arising from deviations in motion patterns.

Let $\mathbf{P}_t$ denote the error covariance matrix at step $t$, and set the measurement matrix to the identity matrix, thereby simplifying the Kalman gain during the prediction phase to
\begin{equation}
    \mathbf{K}_{t+1} = \left( \mathbf{F} \mathbf{P}_t \mathbf{F}^\top + \mathbf{Q} \right) \left( \mathbf{F} \mathbf{P}_t \mathbf{F}^\top + \mathbf{Q} + \mathbf{R} \right)^{-1},
\end{equation}
where $\mathbf{R}$ is measurement noise covariance matrix. Subsequently, the error covariance matrix is updated to
\begin{equation}
    \mathbf{P}_{t+1} = \left( \mathbf{I} - \mathbf{K}_{t+1} \right) \left( \mathbf{F} \mathbf{P}_t \mathbf{F}^\top + \mathbf{Q} \right).
\end{equation}

Let $\boldsymbol{\Sigma}_t$ denote the $3\times3$ upper-left submatrix of the error covariance matrix $\mathbf{P}_t$, which describes the position uncertainty of a dynamic obstacle. It will be used to enable uncertainty-aware dynamic obstacle avoidance.

\subsection{Uncertainty-aware Sector Geometry Modeling}

To achieve precise dynamic obstacle avoidance, we propose an uncertainty-aware sector to represent the potential future occupancy zones of a dynamic obstacle, as illustrated in Fig.~\ref{fig:sector}. The geometry of this sector is adaptively determined by the obstacle's motion state and its associated motion uncertainty. Unlike the existing method~\cite{lu2024fapp}, which uses ellipsoidal modeling and generates overly conservative occupied regions that make optimization prone to failure, the proposed sector modeling method effectively releases the lateral feasible region while fully preserving the high-confidence main direction of motion, thereby enhancing the success rate of trajectory optimization.

Let $\mathbf{v}_i(t)$ denote the projection of the velocity of dynamic object $i$ at time $t$ onto the $x$-$y$ plane, $r_i(t)$ represent the diagonal length of the projection of the bounding box of object i onto the $x$-$y$ plane at time $t$, and $\boldsymbol{\Sigma}_i(t)$ be the top-left $2\times2$ submatrix of the position error covariance matrix of the object $i$ at time $t$. The radius of the sector is determined by the magnitude of the dynamic object's velocity and its position error covariance, defined as
\begin{equation}
    R_i(t) = \Vert \mathbf{v}_i(t) \Vert h + k \sqrt{\hat{\mathbf{v}}_i(t)^\top \boldsymbol{\Sigma}_i(t) \hat{\mathbf{v}}_i(t)} + \dfrac{r_i(t)}{2} + s,
\end{equation}
where $h$ and $k$ are hyperparameters used to balance velocity estimation and position uncertainty, respectively, $\hat{\mathbf{v}}_i(t)$ is the unit direction vector of $\mathbf{v}_i(t)$, and $s$ is the hyperparameter representing the safe distance between robot and obstacles. 

The central angle characterizing lateral uncertainty is
\begin{equation}
    \theta_i(t) = 2\arctan \left( \dfrac{\kappa \sqrt{\hat{\mathbf{u}}_i(t)^\top \boldsymbol{\Sigma}_i(t) \hat{\mathbf{u}}_i(t)}}{\Vert \mathbf{v}_i(t) \Vert h} \right),
\end{equation}
where $\kappa$ is a hyperparameter, $\hat{\mathbf{u}}_i(t)$ is a unit direction vector perpendicular to $\mathbf{v}_i(t)$.

\subsection{Sector-based Dynamic Obstacle Avoidance}

To incorporate uncertainty-aware sectors into trajectory optimization, we design a cost function that is differentiable with respect to both the robot's position and time, enabling the quantification of the collision probability between the robot and dynamic obstacles. Specifically, let $\mathbf{p}(t)$ denote the robot's position at time $t$, and let $\boldsymbol{\mu}_i(t)$ represent the position of dynamic obstacle $i$ at time $t$. The dynamic obstacle avoidance cost consists of radial distance cost $r_i$ and lateral risk cost $l_i$, which are defined as
\begin{equation}
    r_i(\mathbf{p}, t) = R_i(t) - d_i(t),
\end{equation}
\begin{equation}
    l_i(\mathbf{p}, t) = \cos \dfrac{\theta_i(t)}{2} - \cos \left \langle \boldsymbol{\delta}_i(t), \mathbf{v}_i(t) \right \rangle,
\end{equation}
where $\boldsymbol{\delta}_i(t) = \mathbf{p}(t) - \boldsymbol{\mu}_i(t)$ represents the robot's position relative to dynamic obstacle $i$, $d_i(t) = \Vert \boldsymbol{\delta}_i(t) \Vert$ denotes the distance from the robot to dynamic obstacle $i$, and $\langle \mathbf{a}, \mathbf{b} \rangle$ represents the angle between vectors $\mathbf{a}$ and $\mathbf{b}$.

The dynamic obstacle avoidance cost is formulated as
\begin{equation}
    \label{eq:jd}
    J_d(\mathbf{p}, t) = \sum_{i \in C} \sigma_\alpha \left( r_i(\mathbf{p}, t) \right) \cdot \sigma_\beta \left( l_i(\mathbf{p}, t) \right),
\end{equation}
where $C$ denotes the set of dynamic objects, $\alpha$ and $\beta$ are hyperparameters, and $\sigma_\zeta$ is the sigmoid smoothing function used to map values to the interval $(0, 1)$, defined as
\begin{equation}
    \sigma_\zeta(z) = \dfrac{1}{1 + \exp(-\zeta z)}.
\end{equation}
Each term in the summation (\ref{eq:jd}) represents the probability that the robot will collide with a dynamic obstacle. During trajectory optimization, minimizing $J_d$ enables uncertainty-aware avoidance of dynamic obstacles.

\subsection{Uncertainty-aware Trajectory Optimization}

\begin{figure}[t]
    \centering
    \vspace{2 mm}
    \includegraphics[width=\linewidth]{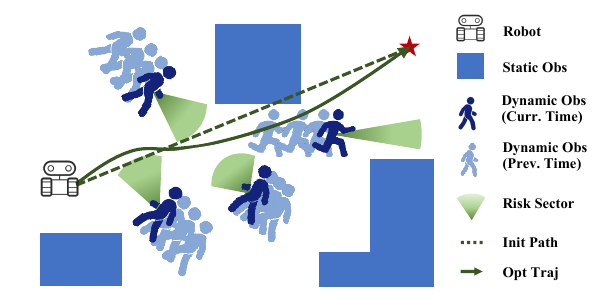}
    \caption{Illustration of uncertainty-aware sectors. Each sector is constructed based on the predicted state and covariance matrix of dynamic obstacles.}
    \vspace{-2 mm}
    \label{fig:sector}
\end{figure}

To obtain a smooth, kinematically feasible, and collision-free trajectory from the starting point to the endpoint, we use the line segment between the two points as the initial path. Then, we uniformly select $N + 1$ points along this path and parameterize these points into a third-order MINCO trajectory~\cite{wang2022geometrically} for optimization, as shown in Fig.~\ref{fig:sector}. The $n$-th segment of the trajectory is represented as
\begin{equation}
    \mathbf{p}_n(t) = \mathbf{C}_n^\top \boldsymbol{\beta}(t), \quad t \in [0, t_n],
\end{equation}
where $\mathbf{C}_n$ contains the coefficients for the trajectory segment, and $\boldsymbol{\beta}(t) = (1, t, t^2, t^3, t^4, t^5)^\top$ is the natural basis.

The uncertainty-aware trajectory optimization problem is formulated as
\begin{subequations}
    \begin{align}
        \underset{\mathbf{C}, \mathbf{t}}{\min} \quad & \sum_{n=1}^{N} \int_{0}^{t_n} \left( \lambda_d J_d(\mathbf{p}_n, t) + \Vert \mathbf{p}_n^{(3)}(t) \Vert^2 + \rho \right) \mathrm{d}t \label{eq:opt}\\
        \mathrm{s.t.} \quad & t_n > 0, \label{eq:time}\\
        \quad & \Xi(\mathbf{p}_n(t)) \ge s, \label{eq:safe}\\
        \quad & \left \Vert  \mathbf{p}_n^{(j)}(t) \right \Vert \le v_j, \quad \forall j \in \lbrace 1, 2 \rbrace,\label{eq:kinematic}\\
        \quad & \mathbf{p}_{n-1}^{(j)}(t_{n-1}) = \mathbf{p}_n^{(j)}(0), \quad \forall j \in \lbrace 0, 1, 2 \rbrace, \label{eq:continue}
    \end{align}
\end{subequations}
where $\mathbf{C} = (\mathbf{C}_1, \mathbf{C}_2, \cdots, \mathbf{C}_n)$, vector $\mathbf{t} = (t_1, t_2, \cdots, t_n)^\top$ denotes the duration of each trajectory segment, $\lambda_d$ is the weight for uncertainty-aware dynamic obstacle avoidance, and $\rho$ is the weight for time regularization. Smoothness is achieved by minimizing the third-order derivatives of the trajectory. Constraint (\ref{eq:time}) ensures that the vector $\mathbf{t}$ lies within the time manifold $\mathbb{R}_+^N$, thereby providing physical meaning. Constraint (\ref{eq:safe}) guarantees that the robot avoids collisions with static obstacles, where $\Xi(\mathbf{p})$ represents the distance from point $\mathbf{p}$ to the nearest static obstacle boundary, which can be computed with linear time complexity~\cite{zhou2019robust}. Constraint (\ref{eq:kinematic}) enforces kinematic feasibility, with $v_1$ and $v_2$ representing the velocity and acceleration limits of the robot, respectively. Constraint (\ref{eq:continue}) ensures the high-order continuity of the segmented trajectory.

To achieve efficient trajectory optimization, we eliminate constraint (\ref{eq:time}) through a differential homeomorphism and construct a time-integral penalty function to address constraints (\ref{eq:safe}) and (\ref{eq:kinematic}), while constraint (\ref{eq:continue}) is inherently satisfied by the properties of the MINCO trajectory. Furthermore, the integrals and derivatives of the higher-order derivatives of the trajectory can be calculated analytically~\cite{wang2022geometrically}. For the uncertainty-aware dynamic obstacle avoidance cost, we calculate it through numerical integration
\begin{equation}
    I_n = \int_{0}^{t_n} J_d(\mathbf{p}_n, t) \mathrm{d}t \approx \sum_{j=0}^{L} \dfrac{t_n}{w_jL} J_d\left( \mathbf{p}_n, \dfrac{jt_n}{L} \right),
\end{equation}
where $\mathbf{w} = (2, 1, 1, \cdots, 1, 2)^\top$, and $L$ represents the number of samples for numerical integration.

The derivative of the numerical integration $I_n$ with respect to coefficients $\mathbf{C}_n$ and time $t_n$ can be calculated as
\begin{equation}
    \dfrac{\partial I_n}{\partial \mathbf{C}_n} = \sum_{j=0}^{L} \dfrac{t_n}{w_jL} \dfrac{\partial J_d}{\partial \mathbf{C}_n},
\end{equation}
\begin{equation}
    \dfrac{\partial I_n}{\partial t_n} = \sum_{j=0}^{L} \left( \dfrac{J_d}{w_jL} + \dfrac{jt_n}{w_jL^2} \dfrac{\partial J_d}{\partial t_n} \right).
\end{equation}
To calculate the gradients of the cost function $J_d$, we present the derivative of the function $\sigma_\zeta$ as
\begin{equation}
    \sigma_\zeta'(z) = \zeta \sigma_\zeta(z) \left( 1 - \sigma_\zeta(z) \right).
\end{equation}
According to the chain rule, for $\boldsymbol{\xi} \in \lbrace \mathbf{C}_n, t_n \rbrace$, the gradient of the cost function $J_d$ can be calculated as
\begin{equation}
    \dfrac{\partial J_d}{\partial \boldsymbol{\xi}} = \sum_{i \in C} \left( \sigma_\beta(l_i) \sigma_\alpha'(r_i) \dfrac{\partial r_i}{\partial \boldsymbol{\xi}} + \sigma_\alpha(r_i) \sigma_\beta'(l_i) \dfrac{\partial l_i}{\partial \boldsymbol{\xi}} \right).
\end{equation}
The derivatives of the radial distance cost $r_i$ is given by
\begin{equation}
    \dfrac{\partial r_i}{\partial \mathbf{C}_n} = -\dfrac{1}{d_i(t)} \boldsymbol{\beta}(t) \boldsymbol{\delta}_i^\top(t),
\end{equation}
\begin{equation}
    \dfrac{\partial r_i}{\partial t_n} = R_i'(t) - \dfrac{\boldsymbol{\delta}_i^\top(t) \boldsymbol{\delta}_i'(t)}{d_i(t)},
\end{equation}
where $R_i'(t)$ can be calculated using the difference approximation, and $\boldsymbol{\delta}_i'(t) = \mathbf{p}'(t) - \mathbf{v}_i(t)$. Similarly, the derivatives of the lateral risk cost $l_i$ is given by
\begin{equation}
    \dfrac{\partial l_i}{\partial \mathbf{C}_n} = - \boldsymbol{\beta}(t) \left( \dfrac{\partial \cos \left \langle \boldsymbol{\delta}_i(t), \mathbf{v}_i(t) \right \rangle}{\partial \mathbf{p}} \right)^\top,
\end{equation}
\begin{equation}
    \dfrac{\partial l_i}{\partial t_n} = - \dfrac{\theta_i'(t)}{2}\sin \dfrac{\theta_i(t)}{2} -\dfrac{\partial \cos \left \langle \boldsymbol{\delta}_i(t), \mathbf{v}_i(t) \right \rangle}{\partial t_n},
\end{equation}
where $\theta_i'(t)$ can be calculated using the difference approximation, and the partial derivatives of $\cos \langle \boldsymbol{\delta}_i(t), \mathbf{v}_i(t) \rangle$ are given in Appendix~\ref{app:derivative} due to their complexity.

Thus, the original problem (\ref{eq:opt}) can be efficiently solved using the gradient-based optimization algorithm~\cite{liu1989limited}.

\section{Experiments and Benchmarks}

\subsection{Implementation Details}

\begin{figure}[h]
    \centering
    \includegraphics[width=\linewidth]{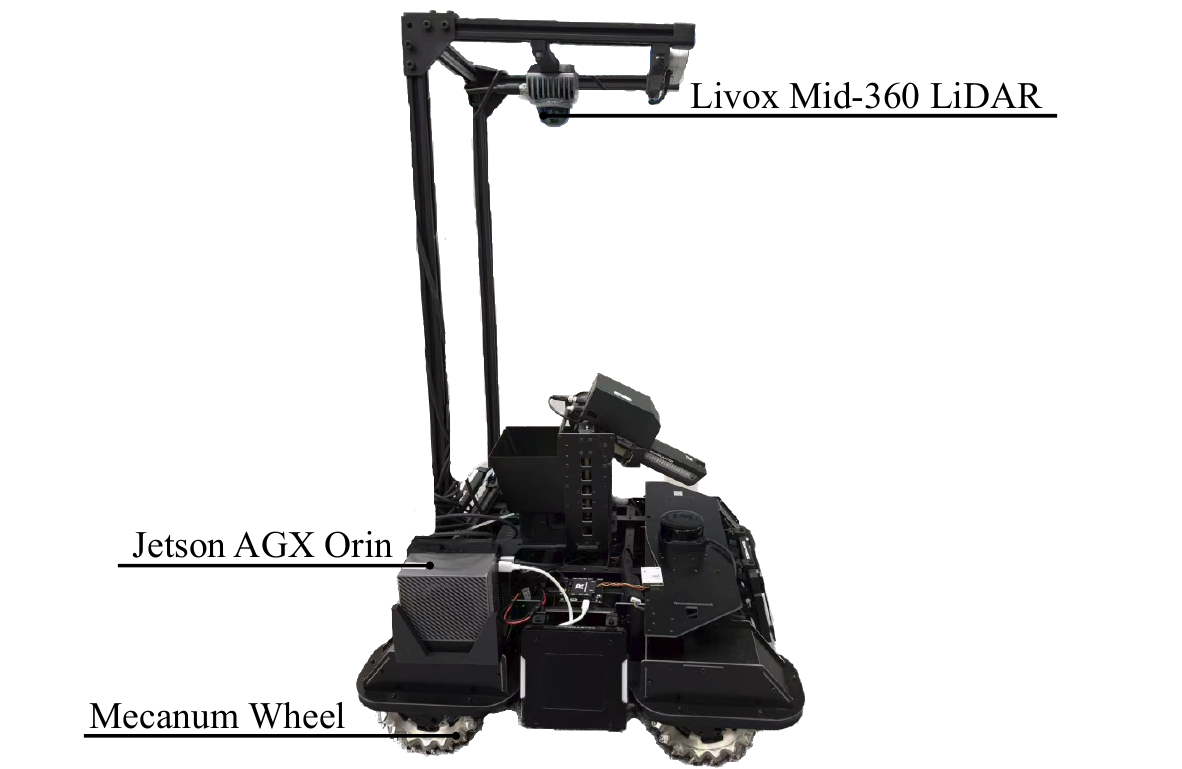}
    \caption{The omnidirectional robot used in real-world experiments. The LiDAR is mounted upside down at the top to fully perceive the robot's surrounding environment. All computations are performed on the onboard Jetson AGX Orin platform.}
    \vspace{-2 mm}
    \label{fig:robot}
\end{figure}

To verify the stability of perception and the effectiveness of uncertainty-aware trajectory optimization in the proposed UCON, we conduct a series of experiments in both simulation and real-world environments, benchmarking it against state-of-the-art methods~\cite{lu2024fapp, xu2025intent}. Real-world experiments are conducted using an omnidirectional robot equipped with Mecanum wheels and a Livox Mid-360 LiDAR, as shown in Fig.~\ref{fig:robot}. The LiDAR features a field of view of $360^\circ \times 59^\circ$ and publishes point clouds at a frequency of $50$ Hz. During the experiment, the robot relies exclusively on this LiDAR for environmental perception. Additionally, the robot is equipped with a Jetson AGX Orin edge computing device, and all simulations and benchmarks are performed on this platform to ensure fairness. The hyperparameters of UCON are set as follows: $\mu_1 = 0.3$, $\mu_2 = 0.3$ and $\mu_3 = 0.4$ for object-level association, $d_0 = 0.6$ and $\gamma = 0.35$ for point-level historical re-association, $h = 0.02$, $k = 2$ and $\kappa = 0.05$ for sector modeling, $\alpha = 15$ and $\beta = 25$ for uncertainty-aware dynamic obstacle avoidance, $\lambda_d = 5 \times 10^3$ to balance optimization terms, and $\rho = 10$ for time regularization.

\subsection{Evaluation of Perception Stability}

\begin{figure}[t]
    \centering
    \vspace{2 mm}
    \includegraphics[width=\linewidth]{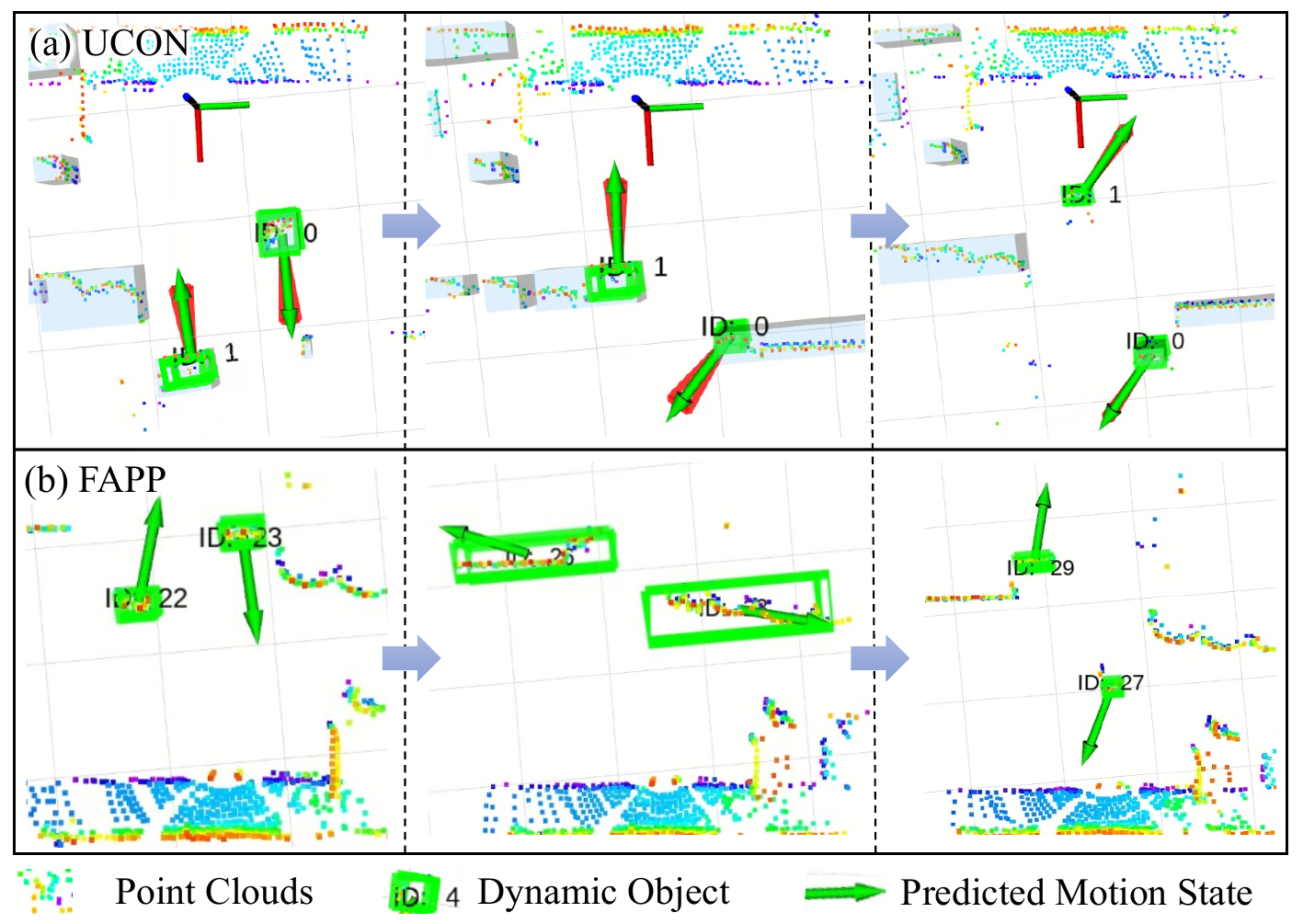}
    \caption{Keyframes from the perception stability benchmark in a real-world environment. (a) Thanks to the proposed point-level historical re-association mechanism, UCON effectively prevents target loss and ensures that the target's identity remains unchanged, thereby greatly mitigating perception instability. (b) The previous method FAPP relies solely on a single-metric object-level association mechanism, frequently resulting in target loss and frequent identity switching, causing perception instability.}
    \vspace{-2 mm}
    \label{fig:perception}
\end{figure}

To verify the ability of the proposed UCON to mitigate perception instability, we benchmark it against state-of-the-art methods in a real-world environment. The comparison includes FAPP~\cite{lu2024fapp}, which is based on LiDAR clustering, and Intent-MPC~\cite{xu2025intent}, which relies on vision and neural networks. The experimental environment consists of several static obstacles and three pedestrians moving at random speeds, with each experiment lasting 243 seconds. Keyframes from the experiments are shown in Fig.~\ref{fig:perception}. We adopt MOTA~\cite{bernardin2006multiple} as the accuracy metric for perception. Additionally, we count the number of identity switching events for the same dynamic object to further evaluate perception stability. During the experiments, the average number of input points per LiDAR scan reaches 4,922, while the UCON achieves an average perception time of 17.78 ms per frame, demonstrating real-time performance. Furthermore, UCON achieves a false negative rate of 4.2\%, a false positive rate of 4.8\%, and a mismatch rate of 5.8\%, resulting in a final MOTA of 85.2\%, as shown in Tab.~\ref{tab:perception}. Due to the complexity of the environment and the interactions among dynamic objects, the compared methods frequently exhibit identity switching. In contrast, UCON effectively manages interactive target segmentation and re-association in dynamic and complex environments, thereby reducing identity switching and achieving stable perception performance.

\begin{table}[t]
    \centering
    \vspace{2 mm}
    \caption{Comparison of Perception Stability}
    \label{tab:perception}
    \begin{tabular}{cccc}
        \toprule
                        & Intent-MPC & FAPP   & \textbf{UCON} (ours) \\
        \midrule
        MOTA            & 66.4\%     & 70.3\% & \textbf{85.2\%}          \\
        Identity Switch & 17         & 16     & \textbf{3}               \\
        \bottomrule
    \end{tabular}
    \vspace{-3 mm}
\end{table}

\subsection{Evaluation of Uncertainty-aware Trajectory Optimization}

To verify the effectiveness of uncertainty-aware trajectory optimization, we benchmark UCON against state-of-the-art methods in a 50 m $\times$ 50 m simulation environment, where 160 static obstacles are randomly placed. Additionally, separate scenarios include $N_d \in \lbrace 50, 80, 110 \rbrace$ dynamic obstacles with random radii (0.2–0.5 m) exhibiting random constant velocity, constant acceleration, or sinusoidal motion. Furthermore, to demonstrate the superiority of the proposed sector-based uncertainty-aware trajectory optimization module, we conduct ablation experiments. 

\begin{figure}[t]
    \centering
    \vspace{2 mm}
    \includegraphics[width=\linewidth]{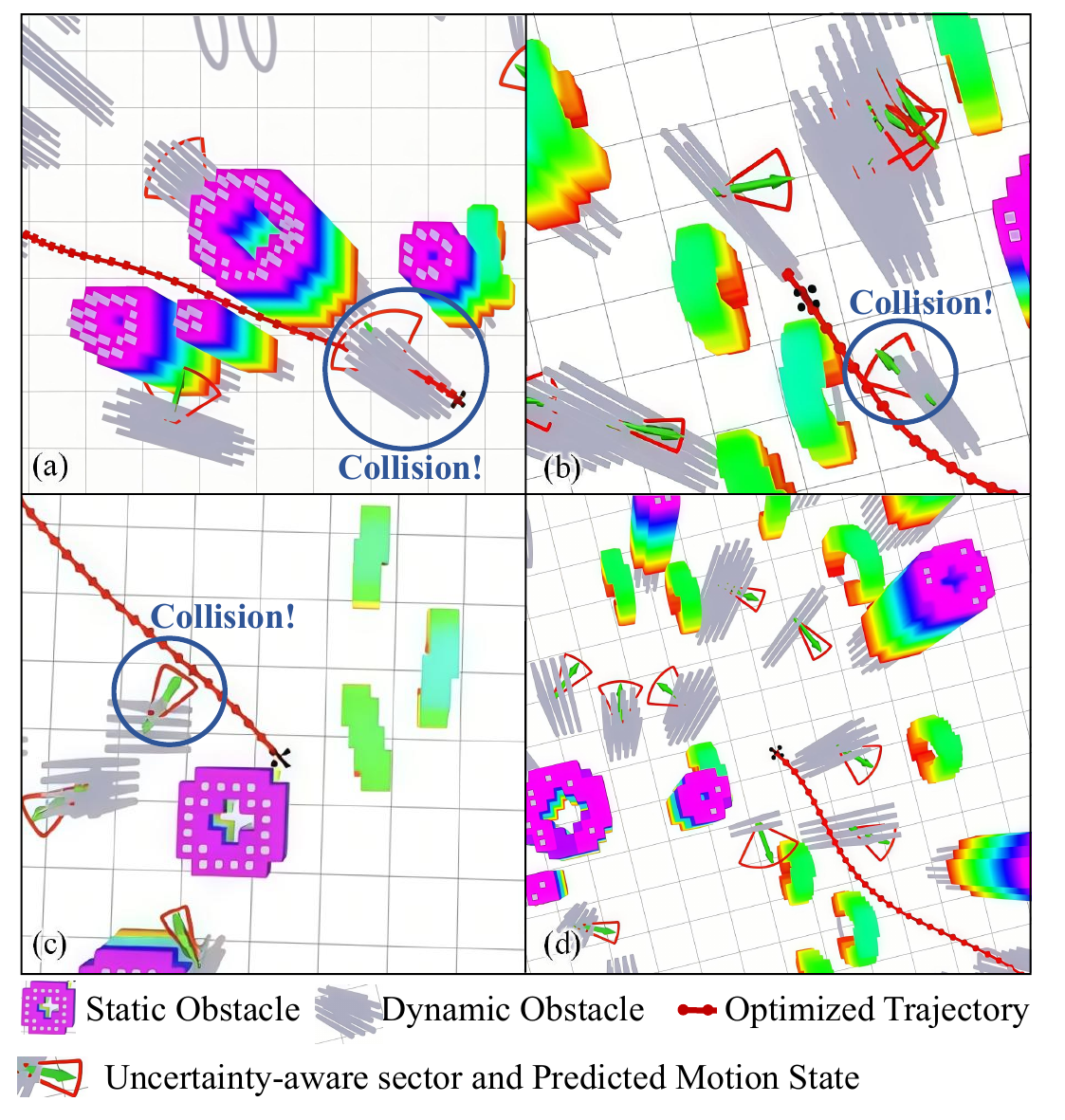}
    \caption{Comparison of trajectories generated in a simulation environment with 80 dynamic obstacles. (a) Due to the incomplete modeling of stability uncertainty for dynamic obstacles, the trajectories generated by FAPP fail to reserve sufficient safety margins, leading to collisions. (b) The trajectory generated by the learning-based Intent-MPC is not ideal due to its low generalization ability. (c) The trajectory generated by the ablation method is unsafe because it does not consider the motion uncertainty of dynamic obstacles. (d) The complete UCON generates a safe and smooth trajectory.}
    \vspace{-3 mm}
    \label{fig:planning}
\end{figure}

Under each dynamic obstacle quantity setting, we independently execute each method fifty times. The trajectories generated by the four methods are visualized in Fig.~\ref{fig:planning}. Thanks to the carefully designed uncertainty-aware dynamic obstacle avoidance objective function, UCON achieves safer trajectories than other methods.  

Furthermore, we summarize the average energy cost $E$ of the trajectory, the average planning time $t_p$, and the planning success rate $\eta$ in Tab.~\ref{tab:planning}, where the average energy cost $E$ is defined as the average jerk over the entire trajectory, and the planning success rate $\eta$ represents the proportion of trajectories that do not collide with any obstacle. The experimental results demonstrate that, compared to other methods, the complete UCON achieves the highest success rate while exhibiting less performance degradation as the number of obstacles increases. Moreover, its planning efficiency and energy cost remain comparable to the best method, supporting the uncertainty-aware trajectory optimization module.

\begin{table}[t]
    \centering
    \vspace{3 mm}
    \caption{Comparison of Trajectory Optimization}
    \label{tab:planning}
    \begin{tabular}{c l c c c}
        \toprule
        Scenario
         & Method          & $E$ (m/s\textsuperscript{3}) & $t_p$ (ms)    & $\eta$ (\%) \\
        \midrule
        \multirow{4}{*}{\makecell[c]{Low Density \\ $N_d = 50$}}
         & Intent-MPC      & 8.01                         & 40.23         & 78          \\
         & FAPP            & 1.39                         & 2.33          & 88          \\
         & UCON (Ablation) & \textbf{1.28}                & \textbf{1.87} & 84          \\
         & UCON (Complete) & 1.57                         & 2.24          & \textbf{90} \\
        \midrule
        \multirow{4}{*}{\makecell[c]{Mid Density \\ $N_d = 80$}}
         & Intent-MPC      & 12.70                        & 58.51         & 62          \\
         & FAPP            & 1.65                         & 2.59          & 78          \\
         & UCON (Ablation) & \textbf{1.40}                & \textbf{2.03} & 74          \\
         & UCON (Complete) & 1.59                         & 2.67          & \textbf{86} \\
        \midrule
        \multirow{4}{*}{\makecell[c]{High Density \\ $N_d = 110$}}
         & Intent-MPC      & 16.93                        & 70.47         & 44          \\
         & FAPP            & 2.06                         & 3.27          & 66          \\
         & UCON (Ablation) & 1.96                         & \textbf{2.85} & 60          \\
         & UCON (Complete) & \textbf{1.78}                & 3.02          & \textbf{74} \\
        \bottomrule
    \end{tabular}
    \vspace{-3 mm}
\end{table} 

\subsection{Navigation in Real-world Dynamic Environments}

\begin{figure}[t]
    \centering
    \vspace{2 mm}
    \includegraphics[width=\linewidth]{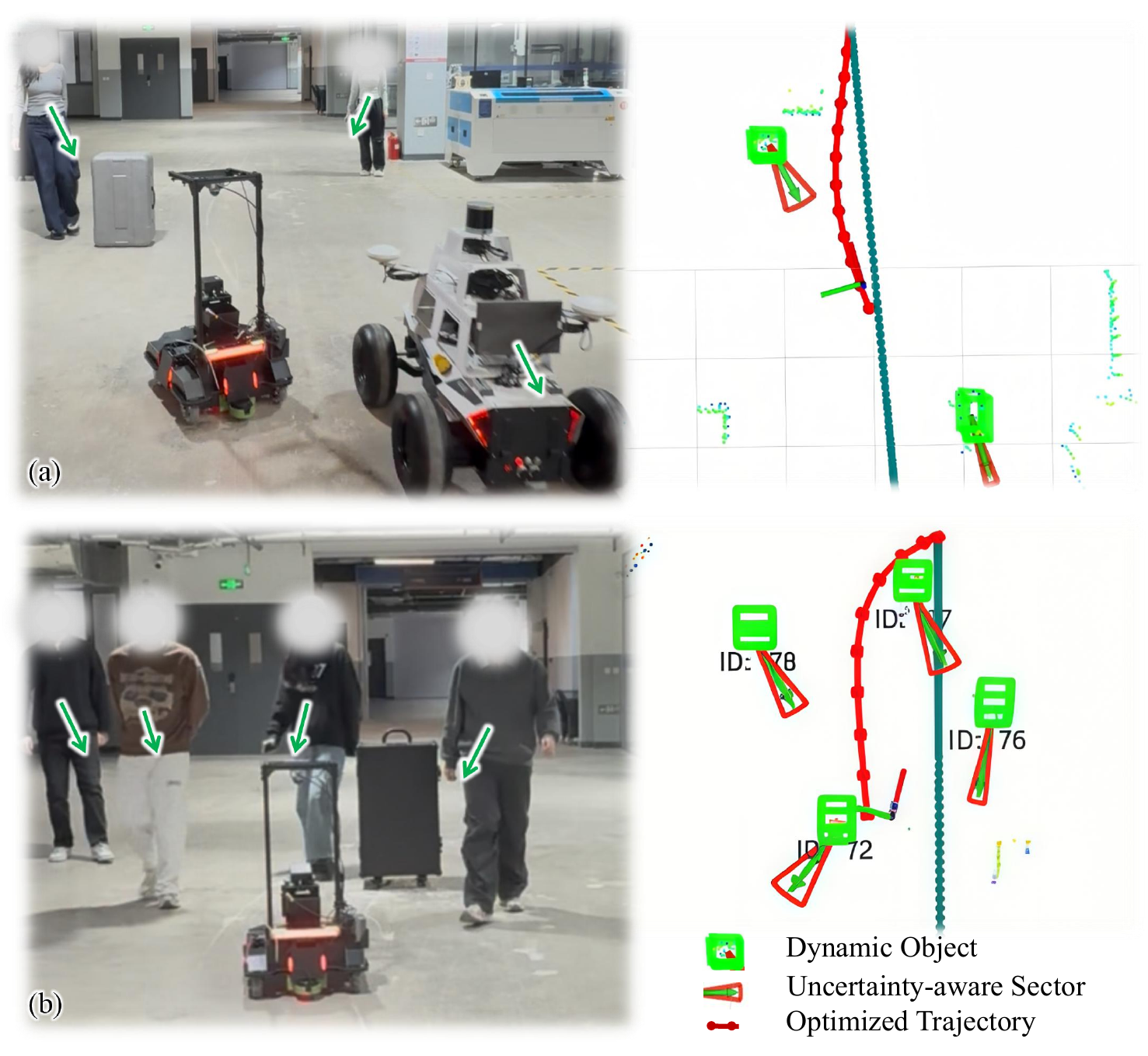}
    \caption{Visualization of real-world experiments. The images on the left illustrate the challenges faced by the robot in experimental environments, while the images on the right display the optimized trajectories of the robot. For privacy protection, visible faces are anonymized in the figures.}
    \vspace{-2 mm}
    \label{fig:realworld}
\end{figure}

To validate the practicality of the proposed UCON, we conduct a series of experiments in real-world dynamic environments. In these scenarios, we employ Faster-LIO~\cite{bai2022faster} for autonomous localization without prior maps. Fig.~\ref{fig:realworld} presents two representative experiments. 

First, we conduct an experiment in an environment that simultaneously contains dynamic and static obstacles with diverse geometries, as shown in Fig.~\ref{fig:realworld}(a). The experimental results demonstrate that UCON can reliably distinguish obstacles of different geometries and generate safe and smooth trajectories, thereby achieving robust dynamic navigation.

Furthermore, as shown in Fig.~\ref{fig:realworld}(b), four volunteers (not affiliated with this work) move simultaneously and intentionally block the robot's trajectory. Thanks to a stable perception algorithm, the proposed UCON successfully identifies and predicts the movement trajectories of these pedestrians. Subsequently, the uncertainty-aware trajectory optimization module constructs risk areas based on the perception results, ultimately generating a collision-free trajectory, enabling the robot to actively avoid pedestrians during navigation.

\section{Conclusions and Limitations}

In this paper, we propose UCON, a unified framework for robust autonomous navigation in dynamic environments, which mitigates perception instability and the uncertainty-optimization mismatch. To mitigate perception instability, we propose a point-level historical re-association mechanism that utilizes historical point clouds to restore identity continuity and improve perceptual stability. To address the uncertainty-optimization mismatch, we propose an uncertainty-aware sector modeling method that integrates anisotropic covariance estimation into a differentiable dynamic obstacle avoidance cost function, thereby establishing a unified pipeline from perception to trajectory optimization. Extensive simulations and real-world experiments demonstrate that UCON improves perception stability and enhances navigation success rates in dense and dynamic scenarios without incurring significant computational overhead.


A limitation of UCON is its omission of semantic-level information, which occasionally causes it to underperform learning-based methods. Future work will address this by incorporating dynamic objects' motion intent.

\appendix

\subsection{Detailed Process of Point-level Historical Re-association}
\label{app:algorithm}

The detailed process of the proposed point-level historical re-association mechanism is outlined below:

\begin{algorithm}[h]
    \SetKwInOut{Input}{Input}
    \SetKwInOut{Output}{Output}
    
    \Input{Unmatched set $\hat C$, Point set $P$, K-D tree $K$}
    \Output{Point sets of re-associated objects $R$}

    Initialize an empty array $R$\;
    \ForEach{$c \in \hat C$}
    {
        $T_c \gets$ ExtractPointCloud($c$)\;
        $N_c \gets |T_c|, \ n \gets 0, \ P_c \gets \varnothing$\;
        \ForEach{$\mathbf{p} \in T_c$}
        {
            $\mathbf{q} \gets$ $K$.FindNearest($\mathbf{p}$, $P$)\;
            \If{$\Vert \mathbf{p} + \mathbf{v}_c \tau - \mathbf{q} \Vert < d_0$}
            {
                $n \gets n + 1$\;
                $P_c \gets P_c \cup \lbrace \mathbf{q} \rbrace$\;
            }
        }
        \If{$n \le \gamma N_c$}
        {
            $P_c \gets \varnothing$\;
        }
        $R$.add($P_c$)\;
    }
\end{algorithm}

\subsection{Detailed Expression of Partial Derivative}
\label{app:derivative}

Let $u_i(t) = \Vert \mathbf{v}_i(t) \Vert$. The cosine of the angle between the vectors $\boldsymbol{\delta}_i(t)$ and $\mathbf{v}_i(t)$ can be calculated by
\begin{equation}
    c_i(\mathbf{p}, t) = \cos \left \langle \boldsymbol{\delta}_i(t), \mathbf{v}_i(t) \right \rangle = \dfrac{\boldsymbol{\delta}_i^\top(t) \mathbf{v}_i(t)}{d_i(t) u_i(t)}.
\end{equation}
According to the chain rule for fractional differentiation, the partial derivative of $c_i(\mathbf{p}, t)$ with respect to position $\mathbf{p}$ is
\begin{equation}
    \dfrac{\partial c_i}{\partial \mathbf{p}} = \dfrac{1}{d_i(t) u_i(t)} \left( \mathbf{v}_i(t) - \dfrac{\boldsymbol{\delta}_i(t) \mathbf{v}_i^\top(t) \boldsymbol{\delta}_i(t)}{\boldsymbol{\delta}_i^\top(t) \boldsymbol{\delta}_i(t)} \right).
\end{equation}
Similarly, the partial derivative of $c_i(\mathbf{p}, t)$ with respect to $t_n$ can be expressed as
\begin{equation}
    \dfrac{\partial c_i}{\partial t_n} = \dfrac{p_i(t) - q_i(t)}{d_i(t) u_i(t)},
\end{equation}
where the components $p_i(t)$ and $q_i(t)$ are defined as
\begin{equation}
    p_i(t) = \boldsymbol{\delta}_i^\top(t) \mathbf{v}_i'(t) + \mathbf{v}_i^\top(t) \boldsymbol{\delta}_i'(t),
\end{equation}
\begin{equation}
    q_i(t) = \boldsymbol{\delta}_i^\top(t) \mathbf{v}_i(t) \left( \dfrac{\boldsymbol{\delta_i}^\top(t) \boldsymbol{\delta_i}'(t)}{\boldsymbol{\delta_i}^\top(t) \boldsymbol{\delta_i}(t)} + \dfrac{\mathbf{v}_i^\top(t) \mathbf{v}_i'(t)}{\mathbf{v}_i^\top(t) \mathbf{v}_i(t)}\right).
\end{equation}
Thus, the derivatives of the lateral risk cost $l_i$ can be derived from these results.


\bibliographystyle{IEEEtran}
\bibliography{reference}

\end{document}